\documentclass[runningheads]{llncs}
\usepackage{graphicx}
\usepackage{graphicx}
\usepackage[usenames, dvipsnames, table]{xcolor}
\usepackage{footnote}
\usepackage{hyperref}[colorlinks=true,allcolors=cblue,citecolor=cblue,linkcolor=cblue,urlcolor=cblue]

 \usepackage[noadjust]{cite} 

\usepackage{textgreek} 
\usepackage{amsmath} 
\usepackage{amsfonts}

\usepackage{float}
\usepackage{placeins} 

\usepackage{tablefootnote} 
\usepackage[roman]{parnotes} 
\usepackage{tabulary}
\usepackage{tabularx}

\usepackage{algorithm,algorithmicx}  
\usepackage[noend]{algpseudocode}
\usepackage[capitalise]{cleveref}
\Crefname{section}{Sec}{Secs.}
\Crefname{figure}{Fig}{Figs.}

\newcommand{\ex}[1]{\textbf{$\triangleright$ Ex.\:}}
\newcommand{\define}[1]{\textbf{$\triangleright$ Definition\:}}
\newcommand{\pradi}{PDR \,}

\usepackage{algorithm,algorithmicx}  
\usepackage[noend]{algpseudocode}
\definecolor{mygray}{gray}{0.5}
\definecolor{cblue}{RGB}{8, 85, 153}
\definecolor{darkblue}{RGB}{1, 43, 112}

\usepackage{changepage} 

\usepackage{adjustbox}
\usepackage{multirow}
\usepackage{booktabs}
\usepackage{tabularx}
\usepackage{placeins}
\usepackage{makecell}

\usepackage{subcaption}

\usepackage{fontawesome}

\newcommand{\norm}[1]{\lVert{#1}\rVert}

\newcommand{\ones}{\mathbf{1}}
\newcommand{\loss}{\mathcal{L}}

\newcommand{\trim}[1]{\textnormal{TRIM}_{f}({#1})}

\newcommand{\bfp}[1]{(\textbf{#1})}

\begin{document}
\title{
Interpreting and improving deep-learning models with reality checks
\thanks{We gratefully acknowledge partial support from 
NSF TRIPODS Grant 1740855, DMS-1613002, 1953191, 2015341, IIS 1741340, ONR grant N00014-17-1-2176, the Center for Science of Information (CSoI), an NSF Science and Technology Center, under grant
agreement CCF-0939370, NSF grant 2023505 on
Collaborative Research: Foundations of Data Science Institute (FODSI),
the NSF and the Simons Foundation for the Collaboration on the Theoretical Foundations of Deep Learning through awards DMS-2031883 and 814639, and a grant from the Weill Neurohub.
}}
\titlerunning{Interpreting deep-learning models with reality checks}
\author{Chandan Singh\inst{1}$^{, *}$ \and
Wooseok Ha\inst{1}$^{, *}$ \and
Bin Yu\inst{1}} 
\authorrunning{C. Singh et al.}
\institute{$^1$University of California, Berkeley, Berkeley CA, USA\\
    \email{\{cs1, haywse, binyu\}@berkeley.edu}\\
    $^*$ Equal contribution \\
}

\maketitle              
\begin{abstract}
Recent deep-learning models have achieved impressive predictive performance by learning complex functions of many variables, often at the cost of interpretability.
This chapter covers recent work aiming to interpret models by attributing importance to features and feature groups for a single prediction.
Importantly, the proposed attributions assign importance to interactions between features, in addition to features in isolation.
These attributions are shown to yield insights across real-world domains, including bio-imaging, cosmology image and natural-language processing.
We then show how these attributions can be used to directly improve the generalization of a neural network or to distill it into a simple model.
Throughout the chapter, we emphasize the use of reality checks to scrutinize the proposed interpretation techniques.\footnote{Code for all methods mentioned in this chapter is available at \href{https://github.com/csinva}{\faGithub\,github.com/csinva} and \href{https://github.com/Yu-Group}{\faGithub\,github.com/Yu-Group}. All methods are implemented in PyTorch~\cite{paszke2017automatic}.}
\end{abstract}

\setcounter{tocdepth}{3}

\section{Interpretability: for what and for whom?}
\label{sec:intro}
Deep neural networks (DNNs) have recently received considerable attention for their ability to accurately predict a wide variety of complex phenomena. However, there is a growing realization that, in addition to predictions, DNNs are capable of producing useful information (i.e. interpretations) about domain relationships contained in data. More precisely, interpretable machine learning can be defined as ``the extraction of \textit{relevant} knowledge from a machine-learning model concerning relationships either contained in data or learned by the model''~\cite{murdoch2019definitions}.\footnote{We include different headings such as explainable AI (XAI), intelligible ML and transparent ML under this definition.}

\begin{figure}[H]
    \centering
    \includegraphics[width=\textwidth]{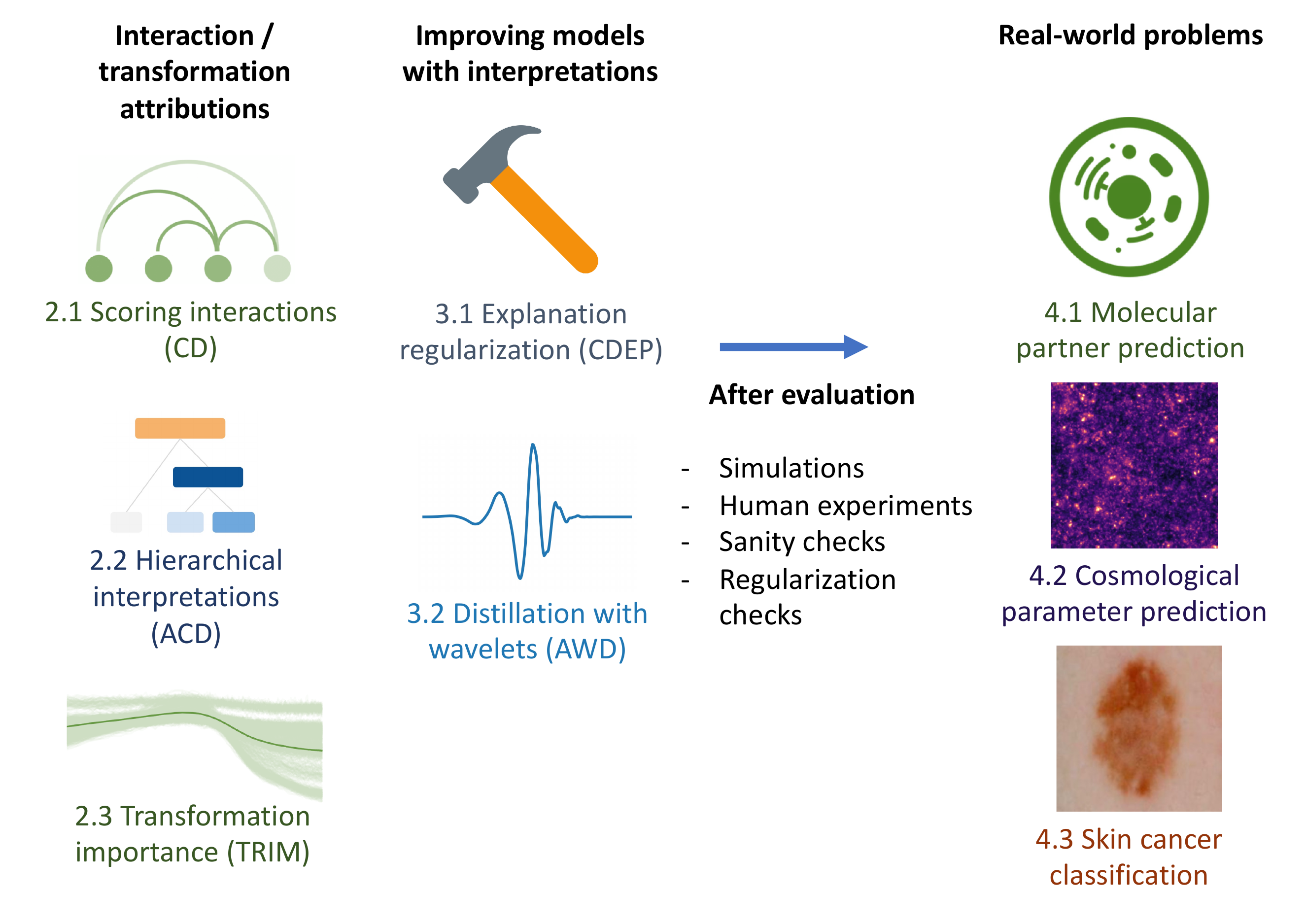}
    \caption{Chapter overview. We begin by defining interpretability and some of its desiderata, following~\cite{murdoch2019definitions} (\cref{sec:intro}). We proceed to overview different methods for computing interpretations for interactions/transformations~(\cref{sec:intro_acd}), including for scoring interactions~\cite{murdoch2018beyond}, generating hierarchical interpretations~\cite{singh2019hierarchical}, and calculating importances for transformations of features~\cite{singh2020transformation}. Next, we show how these interpretations can be used to improve models~(\cref{sec:improve_models_ovw}), including by directly regularizing interpretations~\cite{rieger2020interpretations} and distilling a model through interpretations~\cite{ha2021adaptive}. Finally, we show how these interpretations can be adapted to real-world applications~(\cref{sec:real_data_problems}), including molecular partner prediction, cosmological parameter prediction, and skin-cancer classification.}
    \label{fig:chapter_overview}
\end{figure}

Here, we view knowledge as being \textit{relevant} if it provides insight for a particular audience into a chosen problem.
This definition highlights that interpretability is poorly specified without the context of a particular audience and problem, and should be evaluated with the context in mind. 
This definition also implies that interpretable ML provides correct information (i.e. \textit{knowledge}), and we use the term interpretation, assuming that the interpretation technique at hand has passed some form of \textit{reality check} (i.e. it faithfully captures some notion of reality). 
 
Interpretations have found uses both in their own right, e.g. medicine \cite{litjens2017survey}, policy-making \cite{brennan2013emergence}, and science \cite{angermueller2016deep,vu2018shared}, as well as in auditing predictions themselves in response to issues such as regulatory pressure \cite{goodman2016european} and fairness \cite{dwork2012fairness}.
In these domains, interpretations have been shown to help with evaluating a learned model, providing information to repair a model (if needed), and building trust with domain experts \cite{caruana2015intelligible}.
However, this increasing role, along with the explosion in proposed interpretation techniques~\cite{murdoch2019definitions,olah2017feature,yosinski2015understanding,tsang2017detecting,frosst2017distilling,andreas2016neural,zhang2017interpreting,ha2021adaptive} has raised considerable concerns about the use of interpretation methods in practice~\cite{adebayo2018sanity,gupta2019simple}.
Furthermore, it is unclear how interpretation techniques should be evaluated in the real-world context to advance our understanding of a particular problem.
To do so, we first review some of the desiderata of interpretability, following ~\cite{murdoch2019definitions} among many definitions~\cite{doshi2017roadmap,lipton2016mythos,rudin2018please}, then discuss some methods for critically evaluating interpretations. 

\paragraph{The \pradi desiderata for interpretations}
\label{sec:desiderata}

In general, it is unclear how to select and evaluate interpretation methods for a particular problem and audience.
To help guide this process, we cover the \pradi framework~\cite{murdoch2019definitions}, consisting of three desiderata that should be used to select interpretation methods for a particular problem: predictive accuracy, descriptive accuracy, and relevancy.
\textit{Predictive accuracy} measures the ability of a model to capture underlying relationships in the data (and generally includes different measures of a model's quality of fit)---this can be seen as the most common form of reality check.
In contrast, \textit{descriptive accuracy} measures how well one can approximate what the model has learned using an interpretation method. Descriptive accuracy measures errors during the post-hoc analysis stage of modeling, when interpretations methods are used to analyze a fitted model.
For an interpretation to be trustworthy, one should try to maximize both of the accuracies. In cases where either accuracy is not very high, the resulting interpretations may still be useful.
However, it is especially important to check their trustworthiness through external validation, such as running an additional experiment.
\textit{Relevancy} guides which interpretation to select based on the context of the problem, often playing a key role in determining the trade-off between predictive and descriptive accuracy; however, predictive accuracy and relevancy are not always a trade-off and the examples are shown in~\cref{sec:real_data_problems}.

\paragraph{Evaluating interpretations and additional reality checks}
Techniques striving for interpretations can provide a large amount of fine-grained information, often not just for individual features but also for feature groups~\cite{murdoch2018beyond,singh2019hierarchical}.
As such, it is important to ensure that this added information correctly reflects a model (i.e. has high descriptive accuracy), and can be useful in practice.
This is challenging in general, but there are some promising directions.
One direction, often used in statistical research including causal inference, uses simulation studies to evaluate interpretations.
In this setting, a researcher defines a simple generative process, generates a large amount of data from that process, and trains their statistical or ML model on that data.
Assuming a proper simulation setup, a sufficiently relevant and powerful model to recover the generative process, and sufficiently large training data, the trained model should achieve near-perfect generalization accuracy.
The practitioner then measures whether their interpretations recover aspects of the original generative process. If the simulation captures the reality well, then it can be viewed as a weaker form of reality check.

Going a step further, interpretations can be tested by gathering new data in followup experiments or observations for retrospective validation.
Another direction, which this chapter also focuses on, is to demonstrate the interpretations through domain knowledge which is relevant to a particular domain/audience.
To do so, we closely collaborate with domain experts and showcase how interpretations can inform relevant knowledge in fundamental problems in cosmology and molecular-partner prediction.
We highlight the use of reality checks to evaluate each proposed method in the chapter.


\paragraph{Chapter overview}

A vast line of prior work has focused on assigning importance to individual features, such as pixels in an image or words in a document.
Several methods yield feature-level importance for different architectures. They can be categorized as gradient-based \cite{springenberg2014striving,sundararajan2016gradients,selvaraju2016grad,baehrens2010explain}, decomposition-based \cite{murdoch2017automatic,shrikumar2016not,bach2015pixel}
and others~\cite{dabkowski2017real,fong2017interpretable,ribeiro2016should,zintgraf2017visualizing}, with many similarities among the methods~\cite{ancona2018towards,lundberg2017unified}.
While many methods have been developed to attribute importance to individual features of a model's input, relatively little work has been devoted to understanding interactions between key features.
These interactions are a crucial part of interpreting modern deep-learning models, as they are what enable strong predictive performance on structured data.

Here, we cover a line of work that aims to identify, attribute importance, and utilize interactions in neural networks for interpretation.
We then explore how these attributions can be used to help improve the performance of DNNs.
Despite their strong predictive performance, DNNs sometimes latch onto spurious correlations caused by dataset bias or overfitting \cite{winkler2019association}. As a result, DNNs often exploit bias regarding gender, race, and other sensitive attributes present in training datasets \cite{GargEmbeddings,obermeyer2019dissecting,dressel2018accuracy}. Moreover, DNNs are extremely computationally intensive and difficult to audit.

\cref{fig:chapter_overview} shows an overview of this chapter.
We first overview different methods for computing interpretations~(\cref{sec:intro_acd}), including for scoring interactions~\cite{murdoch2018beyond}, generating hierarchical interpretations~\cite{singh2019hierarchical}, and calculating importances for transformations of features~\cite{singh2020transformation}.
Next, we show how these interpretations can be used to improve models~(\cref{sec:improve_models_ovw}), including by directly regularizing interpretations~\cite{rieger2020interpretations} and distilling a model through interpretations~\cite{ha2021adaptive}. Finally, we show how these interpretations can be adapted to real-world problems~(\cref{sec:real_data_problems}), including molecular partner prediction, cosmological parameter prediction, and skin-cancer classification.

\section{Computing interpretations for feature interactions and transformations}

\label{sec:intro_acd}

This section reviews three recent methods developed to extract the interactions between features that an (already trained) DNN has learned.
First, \cref{subsec:cd_scores} shows how to compute importance scores for groups of features via contextual decomposition (CD), a method which works with LSTMs~\cite{murdoch2018beyond} and arbitrary DNNs, such as CNNs~\cite{singh2019hierarchical}. Next, \cref{subsec:acd} covers agglomerative contextual decomposition (ACD), where a group-level importance measure, in this case CD, is used as a joining metric in an agglomerative clustering procedure. Finally, \cref{subsec:trim} covers transformation importance (TRIM), which allows for computing scores for interactions on transformations of a model's input.
Other methods have been recently developed for understanding model interactions with varying degrees of computational cost and faithfulness to the trained model~\cite{tsang2018can,dhamdhere2019shapley,zhang2020game,wang2021shapley,tsang2017detecting,devlin2019disentangled}.
\subsection{Contextual Decomposition (CD) importance scores for general DNNs}
\label{subsec:cd_scores}

Contextual decomposition breaks up the forward pass of a neural network in order to find an importance score of some subset of the inputs for a particular prediction. For a given DNN $f(x)$, its output is represented as a SoftMax operation applied to logits $g(x)$.  These logits, in turn, are the composition of $L$ layers $g_i$, $i=1,\ldots,L$, such as convolutional operations or ReLU non-linearities:
\begin{align}
f(x) = \text{SoftMax}(g(x)) = \text{SoftMax}(g_L(g_{L-1}(...(g_2(g_1(x)))))).
\end{align}
Given a group of features $\{x_j\}_{j \in S}$, the CD algorithm, $g^{CD}(x)$, decomposes the logits $g(x)$ into a sum of two terms, $\beta(x)$ and $\gamma(x)$. $\beta(x)$ is the importance measure of the feature group $\{x_j\}_{j \in S}$, and $\gamma(x)$ captures contributions to $g(x)$ not included in $\beta(x)$.
\begin{align}
 g^{CD}(x) &  = (\beta(x), \gamma(x)), \\
\beta(x) + \gamma(x) & = g(x).
\end{align}
Computing the CD decomposition for $g(x)$, requires layer-wise CD decompositions $g^{CD}_i(x) = (\beta_i, \gamma_i)$ for each layer $g_i(x)$, where $g_i(x)$ represents the vector of neural activations at the $i$-th layer.
Here, $\beta_i$ corresponds to the importance measure of $\{x_j\}_{j \in S}$ to layer $i$, and $\gamma_i$ corresponds to the contribution of the rest of the input to layer $i$. Maintaining the decomposition requires $\beta_i + \gamma_i = g_i(x)$ for each $i$, the CD scores for the full network are computed by composing these decompositions.
\begin{align}
g^{CD}(x) = g^{CD}_L(g_{L-1}^{CD}(...(g_2^{CD}(g_1^{CD}(x))))).
\end{align}
Note that the above equation shows the CD algorithm $g^{CD}$ takes as input a vector $x$ and for each layer it outputs the pair of vector scores $g^{CD}_i(x) = (\beta_i, \gamma_i)$; and the final output is given by a pair of numbers $g^{CD}(x) = (\beta(x), \gamma(x))$ such that the sum $\beta(x)+\gamma(x)$ equals the logits $g(x)$.

The initial CD work~\cite{murdoch2018beyond} introduced decompositions $g^{CD}_i$ for layers used in LSTMs and the followup work~\cite{singh2019hierarchical} for layers used in CNNs and more generic deep architectures. Below, we give example decompositions for some commonly used layers, such as convolutional layer, linear layer, or ReLU activation.

When $g_i$ is a convolutional or fully connected layer, the layer operation consists of a weight matrix $W$ and a bias vector $b$. The weight matrix can be multiplied with $\beta_{i-1}$ and $\gamma_{i-1}$ individually, but the bias must be partitioned between the two. The bias is partitioned proportionally based on the absolute value of the layer activations. For the convolutional layer, this equation yields only one activation of the output; it must be repeated for each activation.
\begin{align} 
\label{eq:conv_cd}
    \beta_i &= W\beta_{i-1} + \frac{|W\beta_{i-1}|}{|W\beta_{i-1}| + |W\gamma_{i-1}|} \cdot b;  \\
    \gamma_i &= W\gamma_{i-1} + \frac{|W\gamma_{i-1}|}{|W\beta_{i-1}| + |W\gamma_{i-1}|} \cdot b.
\end{align}

Next, for the ReLU activation function,\footnote{See~\cite[Sec. 3.2.2]{murdoch2018beyond} for other activation functions such as sigmoid or hyperbolic tangent.}
importance score $\beta_i$ is computed as the activation of $\beta_{i-1}$ alone and then update $\gamma_i$ by subtracting this from the total activation.
\begin{align} 
\label{eq:relu_cd}
    \beta_{i} &=  \text{ReLU}(\beta_{i-1}); \\ 
    \gamma_{i} &=  \text{ReLU}(\beta_{i-1} + \gamma_{i-1}) - \text{ReLU}(\beta_{i-1}).
\end{align}
For a dropout layer, dropout is simply applied to $\beta_{i-1}$ and $\gamma_{i-1}$ individually.
Computationally, a CD call is comparable to a forward pass through the network $f$.

\subsubsection{Reality check: identifying top-scoring phrases}
When feasible, a common means of scrutinizing what a model has learned is to inspect its most important features and interactions. Table \ref{table:main_top_phrases} 
shows the ACD-top-scoring phrases of different lengths for an LSTM trained on SST (here the phrases are considered from all sentences in the SST's validation set). These phrases were extracted by running ACD separately on each sample in validation set. The score of each phrase was then computed by averaging over the score it received in each occurrence in an ACD hierarchy.
The extracted phrases are clearly reflective of the corresponding sentiment,  providing additional evidence that ACD is able to capture meaningful positive and negative phrases. The paper~\cite{murdoch2018beyond} also shows that CD properly captures negation interactions for phrases.

\begin{table}[H]
    \begin{center}
    \caption{Top-scoring phrases of different lengths extracted by CD on SST's validation set. The positive/negative phrases identified by CD are all indeed positive/negative.}
    \begin{tabular}{l|p{6.8cm}p{6cm}}
    \hline 
    \textbf{Length} &  \textbf{Positive} & \textbf{Negative} \\
    \hline
    1  & pleasurable, glorious & nowhere, grotesque, sleep \\
    \hline
    3 & amazing accomplishment., great fun. & bleak and desperate, conspicuously lacks. \\
    \hline
    5 & a pretty amazing accomplishment. & ultimately a pointless endeavour. \\
    \hline 
    \end{tabular}
    \label{table:main_top_phrases}
    \end{center}
\end{table}

\subsection{Agglomerative Contextual Decomposition (ACD)}
\label{subsec:acd}

Next, we cover agglomerative contextual decomposition (ACD), a general technique that can be applied to a wide range of DNN architectures and data types. Given a prediction from a trained DNN, ACD produces a hierarchical clustering of the input features, along with the contribution of each cluster to the final prediction. This hierarchy is designed to identify clusters of features that the DNN learned are predictive. Throughout this subsection, we use the term CD interaction score between two groups of features to mean the difference between the scores of the combined group and the original groups.

Given the generalized CD scores introduced above, we now introduce the clustering procedure used to produce ACD interpretations.
At a high level, this method is equivalent to agglomerative hierarchical clustering, where the CD interaction score  
is used as the joining metric to determine which clusters to join at each step.
This procedure builds the hierarchy by starting with individual features and iteratively combining them based on the highest interaction scores provided by CD.
The displayed ACD interpretation is the hierarchy, along with the CD importance score at each node.

\begin{figure}[t]
    \centering
    \includegraphics[width=0.85\textwidth]{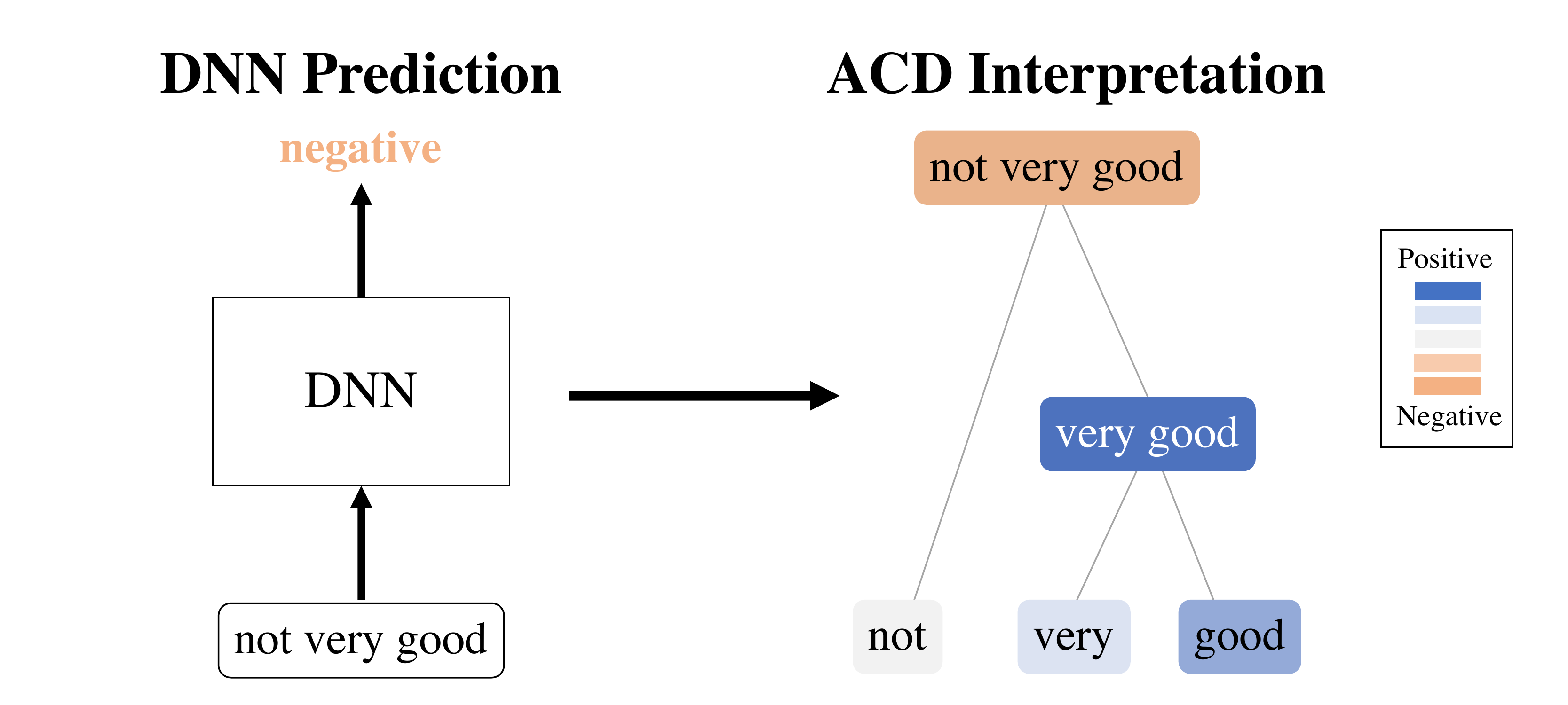} 
    \caption{ACD illustrated through the toy example of predicting the phrase ``not very good'' as negative. Given the network and prediction, ACD constructs a hierarchy of meaningful phrases and provides importance scores for each identified phrase. In this example, ACD identifies that ``very'' modifies ``good'' to become the very positive phrase ``very good'', which is subsequently negated by "not" to produce the negative phrase ``not very good''.}
    \label{fig:intro_acd}
\end{figure}

The clustering procedure proceeds as follows. After initializing by computing the CD scores of each feature individually, the algorithm iteratively selects all groups of features within k\% of the highest-scoring group (where $k$ is a hyperparameter) and adds them to the hierarchy. Each time a new group is added to the hierarchy, a corresponding set of candidate groups is generated by adding individual contiguous features to the original group. For text, the candidate groups correspond to adding one adjacent word onto the current phrase, and for images adding any adjacent pixel onto the current image patch. Candidate groups are ranked according to the CD interaction score, which is the difference between the score of the candidate and the original groups.

\subsubsection{Reality check: human experiment}
\label{sec:exp_details}

Human experiments show that ACD allows users to better reason about the accuracy of DNNs.
Each subject was asked to fill out a survey asking whether, using ACD, they could identify the more accurate of two models across three datasets (SST~\cite{socher2013recursive}, MNIST~\cite{lecun1998mnist} and ImageNet~\cite{imagenet_cvpr09}), and ACD was compared against three baselines: CD \cite{murdoch2018beyond}, Integrated Gradients (IG) \cite{sundararajan2016gradients}, and occlusion \cite{li2016understanding,zeiler2014visualizing}.
Each model uses a standard architecture that achieves high classification accuracy, and has an analogous model with substantially poorer performance obtained by randomizing some fraction of its weights while keeping the same predicted label. 
The objective of this experiment was to determine if subjects could use a small number of interpretations produced by ACD to identify the more accurate of the two models.

For each question, $11$
subjects were given interpretations from two different models (one high-performing and one with randomized weights), and asked to identify which of the two models had a higher generalization accuracy.
To prevent subjects from simply selecting the model that predicts more accurately for the given example, for each question a subject is shown two sets of examples: one where only the first model predicts correctly and one where only the second model predicts correctly (although one model generalizes to \textit{new} examples much better).



\begin{figure}[hbtp]
    \centering
    \includegraphics[width=0.5\textwidth]{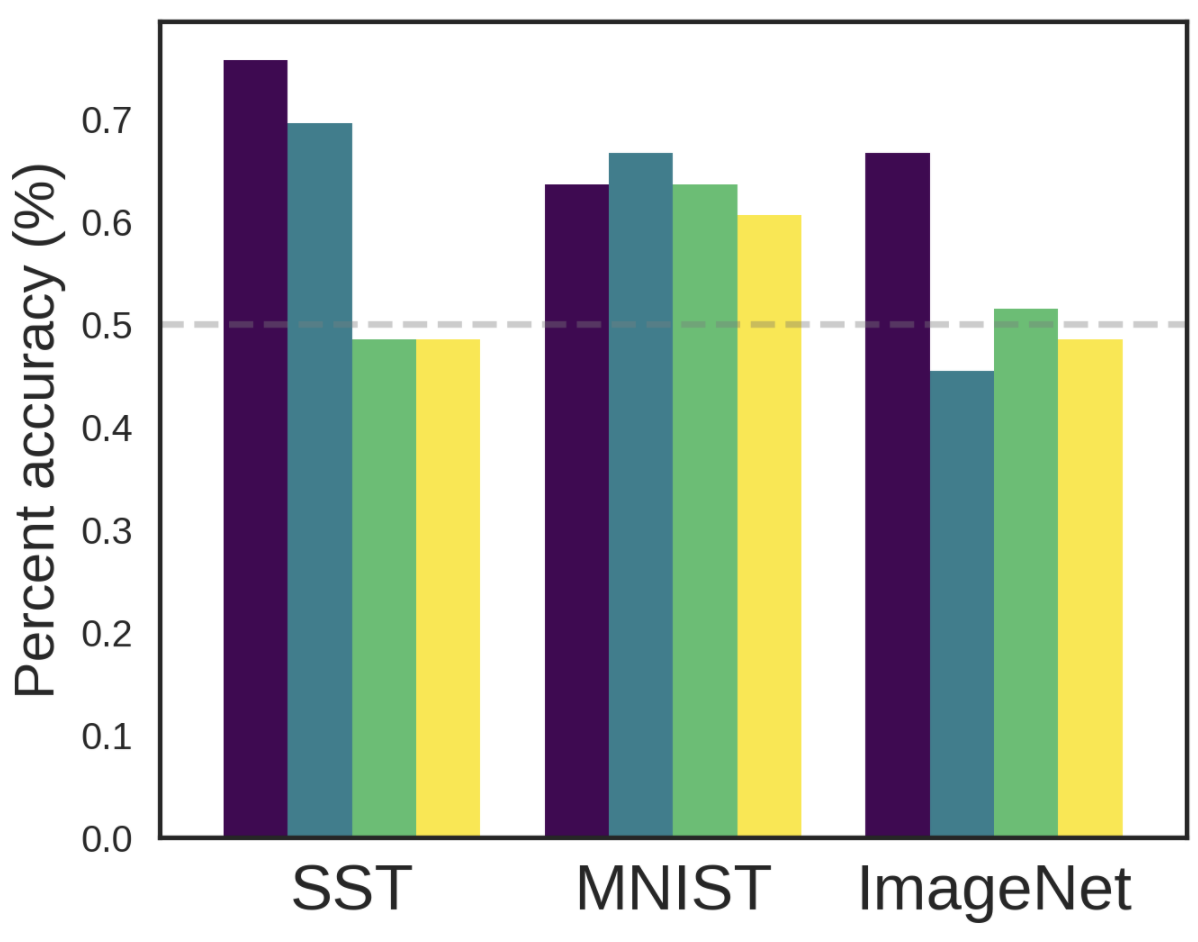}
    \includegraphics[width=0.2\textwidth]{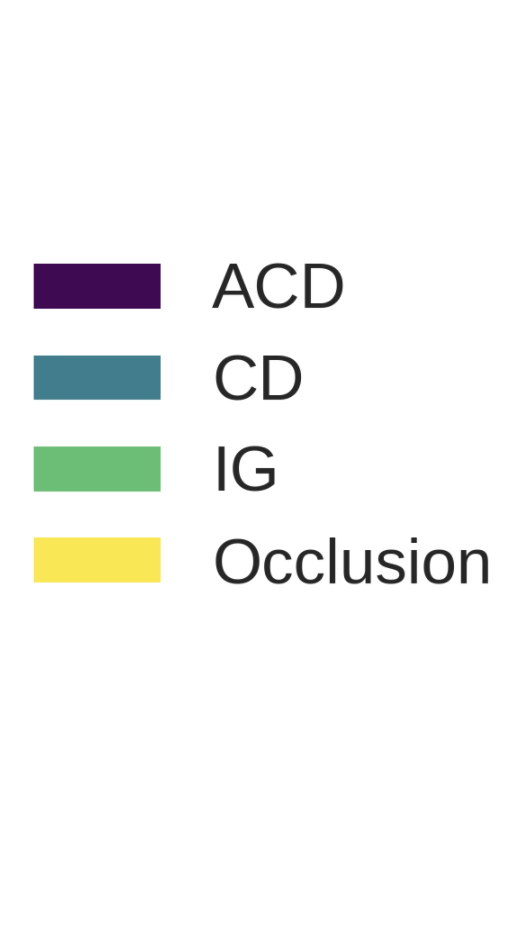}
    \caption{Results for human studies. Binary accuracy for whether a subject correctly selected the more accurate model using different interpretation techniques.}
    \label{fig:human_results}
\end{figure}


Fig~\ref{fig:human_results} shows the results of the survey. For SST, humans were better able to identify the strongly predictive model using ACD compared to other baselines, with only ACD and CD outperforming random selection (50\%). Based on a one-sided two-sample t-test, the gaps between ACD and IG/Occlusion are significant, but not the gap between ACD and CD. In the simple setting of MNIST, ACD performs similarly to other methods. When applied to ImageNet, a more complex dataset, ACD substantially outperforms prior, non-hierarchical methods, and is the only method to outperform random chance. The paper~\cite{singh2019hierarchical} also contains results showing that the ACD hierarchy is robust to adversarial perturbations.

\subsection{Transformation importance with applications to cosmology (TRIM)}
\label{subsec:trim}

Both CD and ACD show how to attribute importance to interactions between features. However, in many cases, raw features such as pixels in an image or words in a document may not be the most meaningful spaces to perform interpretation. When features are highly correlated or features in isolation are not semantically meaningful, the resulting attributions need to be improved.

\begin{figure}[hbtp]
    \centering
    \includegraphics[width=0.7\textwidth]{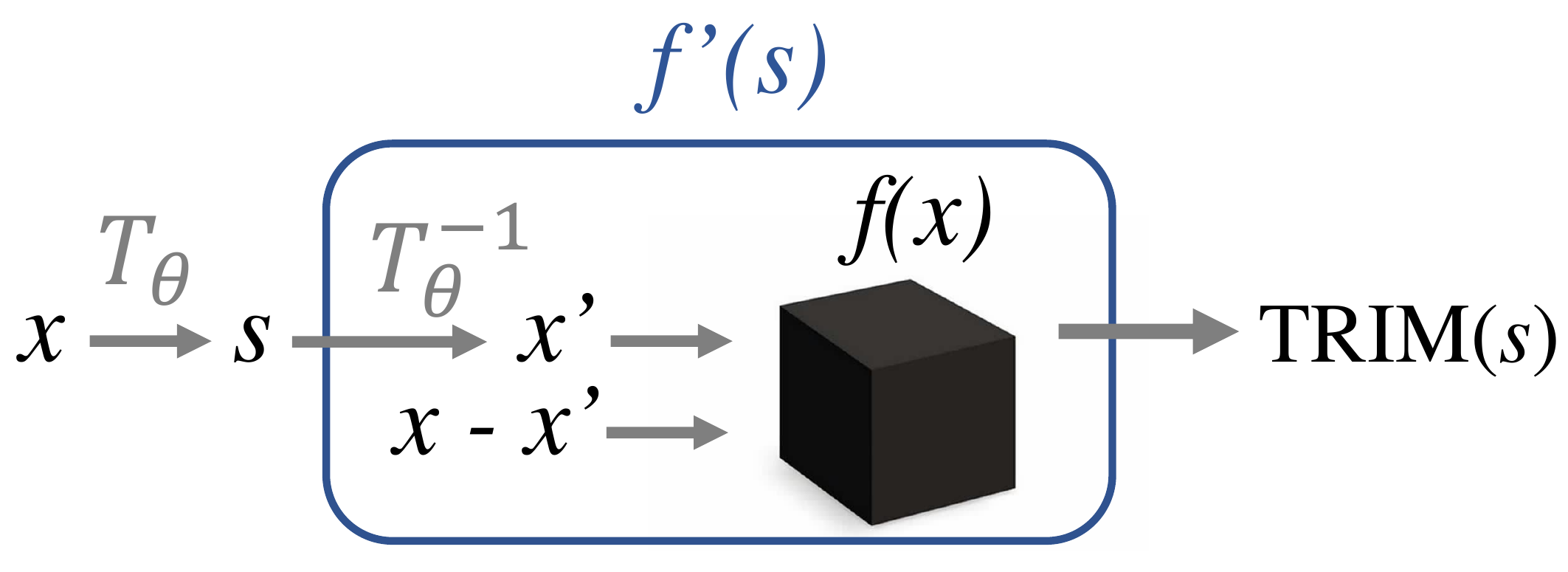}
    \caption{TRIM: Attributing importance to a transformation of an input $T_\theta(x)$ given a model $f(x)$.}
    \label{fig:t}
\end{figure}

To meet this challenge, TRIM (\underline{Tr}ansformation \underline{Im}portance) attributes importance to transformations of the input features (see \cref{fig:t}).
This is critical for making interpretations relevant to a particular audience/problem, as attributions in a domain-specific feature space (e.g. frequencies or principal components) can often be far more interpretable than attributions in the raw feature space (e.g. pixels or biological readings). Moreover, features after transformation can be more independent, semantically meaningful, and comparable across data points.
The work here focuses on combining TRIM with CD, although TRIM can be combined with any local interpretation method.

TRIM aims to interpret the prediction made by a model $f$ given a single input $x$.
The input $x$ is in some domain $\mathcal X$, but we desire an explanation for its representation $s$ in a different domain $\mathcal S$, defined by a mapping $T: \mathcal X \to \mathcal S$, such that $s = T (x)$. 
For example, if $x$ is an image, $s$ may be its Fourier representation, and $T$ would be the Fourier transform. Notably, this process is entirely post-hoc: the model $f$ is already fully trained on the domain $\mathcal X$. By reparametrizing the network as shown in \cref{fig:t}, we can obtain attributions in the domain $\mathcal S$. If we require that the mapping $T$ be invertible, so that $x = T^{-1} (s)$, we can represent each data point $x$ with its counterpart $s$ in the desired domain, and the function to interpret becomes $f' = f \circ T^{-1}$; the function $f'$ can be interpreted with any existing local interpretation method $attr$ (e.g. LIME \cite{ribeiro2016should} or CD~\cite{murdoch2018beyond,singh2019hierarchical})). Note that if the transformation $T$ is not perfectly invertible (i.e. $x \neq x'$), then the residuals $x-x'$ may also be required for local interpretation. For example, they are required for any gradient-based attribution method to aid in computing $\partial f' / \partial s$.\footnote{If the residual is not added, the gradient of $f'=f\circ T^{-1}$ requires $\partial f/\partial x \vert_{x'}$, which can potentially cause evaluation of $f$ at the out-of-distribution examples $x'\neq x$.}
Once we have the reparameterized function $f'(s)$, we need only specify which part of the input to interpret to define TRIM:

\begin{definition} Given a model $f$, an input $x$, a mask $M$, a transformation $T$, and an attribution method $attr$, 
\label{def_t}
\vspace{2pt}
\begin{align*}
    \textup{TRIM}(s) = attr\left(f'; s \right) \\
    \textup{where } f' = f \circ T^{-1}, s = M \odot T(x) 
\end{align*}
\vspace{2pt}
Here $M$ is a mask used to specify which parts of the transformed space to interpret and $\odot$ denotes elementwise multiplication.
\end{definition}

In the work here, the choice of attribution method $attr$ is CD, and $attr\left(f; x', x \right)$ represents the CD score for the features $x'$ as part of the input $x$.
This formulation does not require that $x'$ simply be a binary masked version of $x$; rather, the selection of the mask $M$ allows a human/domain scientist to decide which transformed features to score. In the case of image classification, rather than simply scoring a pixel, one may score the contribution of a frequency band to the prediction $f(x)$. This general setup allows for attributing importance to a wide array of transformations. For example, $T$ could be any invertible transform (e.g. a wavelet transform), or a linear projection (e.g. onto a sparse dictionary). Moreover, we can parameterize the transformation $T_\theta$ and learn the parameters $\theta$ to produce a desirable representation (e.g. sparse or disentangled).

\begin{figure}[hbtp]
    \centering
        \includegraphics[width=0.9\textwidth]{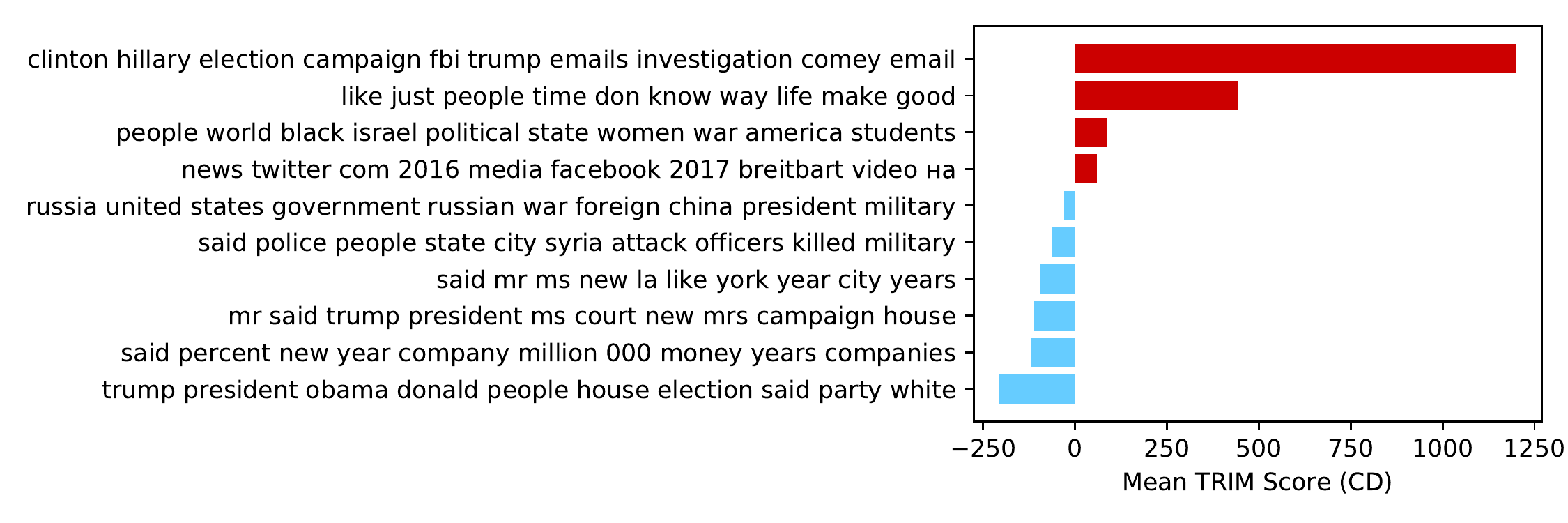}
        
    \caption{TRIM attributions for a fake-news classifier based on a topic model transformation. Each row shows one topic, labeled with the top ten words in that topic. Higher attributions correspond to higher contribution to the class \textit{fake}. Calculated over all points which were accurately classified as \textit{fake} in the test set (4,160 points).}
    \label{fig:fakenews_example}
    
\end{figure}

As a simple example, we investigate a text-classification setting using TRIM. We train a 3-layer fully connected DNN with ReLU activations on the Kaggle Fake News dataset,\footnote{\url{https://www.kaggle.com/c/fake-news/overview}} achieving a test accuracy of 94.8\%. The model is trained directly on a bag-of words representation, but TRIM can provide a more succinct space via a topic model transformation (learned via latent dirichlet allocation \cite{blei2003latent}). \cref{fig:fakenews_example} shows the mean attributions for different topics when the model predicts \textit{Fake}. Interestingly, the topic with the highest mean attribution contains recognizable words such as \textit{clinton} and \textit{emails}.

\subsubsection{Simulation}\footnote{While simulation study in general is not reality check, as we mentioned in~\cref{sec:intro}, it can be seen as a weaker form of reality check as long as it captures the reality.}


In the case of a perfectly invertible transformation, such as the Fourier transform, TRIM simply measures the ability of the underlying attribution method (in this case CD) to correctly attribute importance in the transformed space. We run synthetic simulations showing the ability of TRIM with CD to recover known groundtruth feature importances. Features are generated i.i.d. from a standard normal distribution. Then, a binary classification outcome is defined by selecting a random frequency and testing whether that frequency is greater than its median value. Finally, we train a 3-layer fully connected DNN with ReLU activations on this task and then test the ability of different methods to assign this frequency the highest importance. \cref{tab:simulations} shows the percentage of errors made by different methods in such a setup. CD has the lowest error on average, compared to popular baselines.


\begin{table}[H]

\centering
\small
    \caption{Error (\%) in recovering a groundtruth important frequency in simulated data using different attribution methods with TRIM, averaged over 500 simulated datasets.}
\begin{tabular}{cccc}
    \hline
    CD  & DeepLift \cite{shrikumar2016not} & SHAP \cite{lundberg2017unified} & Integrated Gradients \cite{sundararajan2016gradients}\\
    \hline
    \textbf{0.4 $\pm$ 0.282}  & 3.6 $\pm$ 0.833     & 4.0 $\pm$ 0.897  & 4.2 $\pm$ 0.876    \\
    \hline
\end{tabular}
\label{tab:simulations}
\end{table}

\section{Using attributions to improve models}
\label{sec:improve_models_ovw}


This section shows two methods for using the attributions introduced in \cref{sec:intro_acd} to directly improve DNNs.
\cref{sec:intro_cdep} shows how CD scores can be penalized during training to improve generalization in interesting ways and \cref{subsec:wavelets} shows how attribution scores can be used to distill a DNN into a simple data-driven wavelet model.

\subsection{Penalizing explanations to align neural networks with prior knowledge (CDEP)}
\label{sec:intro_cdep}

While much work has been put into developing methods for explaining DNNs, relatively little work has explored the potential to use these explanations to help build a better model.
Some recent work proposes forcing models to attend to certain regions~\cite{burns2018women,mitsuhara2019embedding,du2019learning} or penalizing the gradients or expected gradients of a neural network~\cite{ross2017right,bao2018deriving,du2019learning,ross2018improving,liu2019incorporating,erion2019learning}.

Here, we cover \underline{c}ontextual \underline{d}ecomposition \underline{e}xplanation \underline{p}enalization (CDEP), a method which leverages CD to enable the insertion of domain knowledge into a model~\cite{rieger2020interpretations}. Given prior knowledge in the form of importance scores, CDEP works by allowing the user to directly penalize importances of certain features or feature interactions. This forces the DNN to not only produce the correct prediction, but also the correct explanation for that prediction. CDEP can be applied to arbitrary DNN architectures and is often orders of magnitude faster and more memory efficient than recent gradient-based methods~\cite{ross2017right,erion2019learning}; CDEP offers significant computational improvements, since, unlike gradient-based attributions, the CD score is computed along the forward pass, only first derivatives are required for optimization, early layers can be frozen, and all activations of a DNN do not need to be cached to perform backpropagation; furthermore, with gradient-based methods the training requires the storage of activations and gradients for all layers of the network as well as the gradient with respect to the input, whereas penalizing CD requires only a small constant amount of memory more than standard training.

CDEP works by augmenting the traditional objective function used to train a neural network, as displayed in \cref{eq:general_method} with an additional component. In addition to the standard prediction loss $\mathcal{L}$, which teaches the model to produce the correct predictions by penalizing wrong predictions, we  add an explanation error $\mathcal{L}_{\text{expl}}$, which teaches the model to produce the correct explanations for its predictions by penalizing wrong explanations.  In place of the prediction and labels $f_\theta(X), y$, used in the prediction error $\mathcal{L}$, the explanation error $\mathcal{L}_{\text{expl}}$ uses the explanations produced by an interpretation method $\text{expl}_\theta(X)$, along with targets provided by the user $\text{expl}_X$. The two losses are weighted by a hyperparameter $\lambda \in \mathbb{R}$:

\begin{equation}
\begin{split}
\label{eq:general_method}
\hat{\theta} = \underset{\theta}{\text{argmin}} \: \overbrace{\mathcal{L}\left(f_\theta(X), y\right)}^{\text{Prediction error}}
+ \lambda \overbrace{\mathcal{L}_{\text{expl}}\left(\text{expl}_\theta(X), \text{expl}_X\right)}^{\text{Explanation error}}
\end{split}
\end{equation}

CDEP uses CD as the explanation function used to compute $\text{expl}_\theta(X)$, allowing the penalization of interactions between features.
We now substitute the above CD scores into the generic equation in \cref{eq:general_method} to arrive at CDEP as it is used in this chapter. 
We collect from the user, for each input $x_i$, a collection of feature groups $x_{i, S}$, $x_i \in \mathbb{R}^d, S \subseteq \{1,...,d\}$, along with explanation target values $\text{expl}_{x_{i, S}}$, and use the $\|\cdot \|_1$ loss for $\mathcal{L}_{\text{expl}}$. This yields a vector $\beta(x_j)$ for any subset of features in an input $x_j$ which we would like to penalize. We can then collect prior knowledge label explanations for this subset of features, $\text{expl}_{x_j}$ and use it to regularize the explanation: 

\begin{equation}
\hat{\theta} = \underset{\theta}{\text{argmin}} \: \overbrace{ \underset{i}{\sum} \underset{c}{\sum} -y_{i, c} \log f_{\theta}(x_i)_{c}}^{\text{Prediction error}} 
 +  \lambda \overbrace{\sum_i\sum_S  ||\beta(x_{i, S}) - \text{expl}_{x_{i, S}}||_1}^{\text{Explanation error}}
\label{eq:specific_methods}
\end{equation}

In the above, $i$ indexes each individual example in the dataset, $S$ indexes a subset of the features for which we penalize their explanations, and $c$ sums over each class. 


The choice of prior knowledge 
explanations $\text{expl}_X$ is dependent on the application and the existing domain knowledge. 
CDEP allows for penalizing arbitrary interactions between features, allowing the incorporation of a very broad set of domain knowledge. In the simplest setting, practitioners may precisely provide prior knowledge
human explanations for each data point.
To avoid assigning human labels, one may utilize programmatic rules to identify and assign prior knowledge 
importance to regions, which are then used to help the model identify important/unimportant regions.
In a more general case, one may specify importances of different feature interactions.

\subsubsection{Towards reality check: ColorMNIST task}
\label{subsec:mnist}

\begin{figure*}[htb!]
	\begin{center}
		\includegraphics[width=\linewidth]{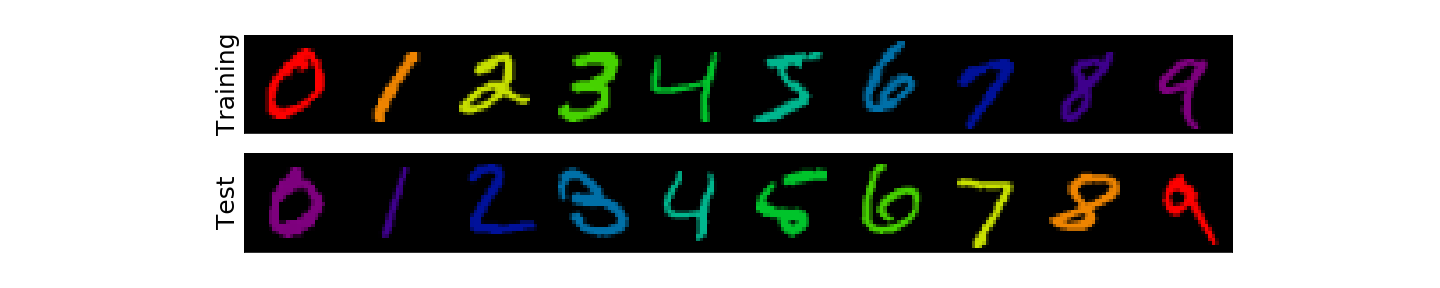}
	\end{center}
	
	\caption{ColorMNIST: the shapes remain the same between the training set and the test set, but the colors are inverted.}
	\label{fig:colorMNIST}
\end{figure*}

Here, we highlight CDEP's ability to alter which features a DNN uses to perform digit classification.
Similar to one previous study \cite{li2019repair}, we alter the MNIST dataset to include three color channels and assign each class a distinct color, as shown in \cref{fig:colorMNIST}. An unpenalized DNN trained on this biased data will completely misclassify a test set with inverted colors, dropping to 0\% accuracy (see \cref{tab:mnist}), suggesting that it learns to classify using the colors of the digits rather than their shape. 

Interestingly, this task can be approached  by minimizing the contribution of pixels in isolation (which only represent color) while maximizing the importance of groups of pixels (which can represent shapes). To do this, CDEP penalizes the CD contribution of sampled single-pixel values, following \cref{eq:specific_methods}. Minimizing the contribution of single pixels encourages the DNN to focus instead on groups of pixels.
\cref{tab:mnist} shows that CDEP can partially divert the network's focus on color to also focus on digit shape. The table includes 2 baselines: penalization of the squared gradients (RRR) \cite{ross2017right} and Expected Gradients (EG) \cite{erion2019learning}.
The baselines do not improve the test accuracy of the model on this task above the random baseline, while CDEP significantly improves the accuracy to 31.0\%.
\begin{table}[H]
    \caption{Test Accuracy on ColorMNIST. CDEP is the only method that captures and removes color bias. All values averaged over thirty runs. Predicting at random yields a test accuracy of 10\%.} 
    \label{tab:mnist}
    \begin{center}
    \begin{small}
     \begin{tabular}{lccccc}
     \toprule
     & Vanilla & CDEP & RRR & Expected Gradients \\
     \midrule
     ColorMNIST & 0.2 $\pm$ 0.2 & \textbf{31.0 $\pm$ 2.3} & 0.2 $\pm$ 0.1 & 10.0 $\pm$ 0.1 \\
     \bottomrule
     \end{tabular}
    \end{small}
    \end{center}
\end{table}

The paper~\cite{rieger2020interpretations} further shows how CDEP can be applied to diverse applications, such as notions of fairness in the COMPAS dataset~\cite{larson2016we} and in natural-language processing.
\subsection{Distilling adaptive wavelets from neural networks with interpretations}
\label{subsec:wavelets}

One promising approach to acquiring highly predictive interpretable models is model distillation. Model distillation is a technique which distills the knowledge in one model into another model. Here, we focus on the case where we distill a DNN into a simple, wavelet model. Wavelets have many useful properties, including fast computation, an orthonormal basis, and interpretation in both spatial and frequency domains~\cite{mallat2008wavelet}. Here, we cover \underline{a}daptive \underline{w}avelet \underline{d}istillation (AWD), a method to learn a valid wavelet by distilling information from a trained DNN~\cite{ha2021adaptive}. 

\cref{eq:awd} shows the three terms in the formulation of the method. $x_i$ represents the $i$-th input signal, $\widehat{x}_i$ represents the reconstruction of $x_i$, $h$ and $g$ represent the lowpass and highpass wavelet filters, and $\Psi x_i$ denotes the wavelet coefficients of $x_i$. $\lambda$ is a hyperparameter penalizing the sparsity of the wavelet coefficients, which can help to learn a compact representation of the input signal and $\gamma$ is a hyperparameter controlling the strength of the interpretation loss, which controls how much to use the information coming from a trained model $f$:

\begin{equation}
    \label{eq:awd}
    \underset{h,g}{\text { minimize }}\loss(h,g)= \underbrace{\frac{1}{m}\sum_{i}\norm{x_i - \widehat{x}_i}_{2}^{2}}_{\text {Reconstruction loss }}
    +\underbrace{\frac{1}{m}\sum_i W(h, g, x_i; \lambda)}_{\text {Wavelet loss }}
    +\underbrace{\gamma \sum_{i} \norm{\trim{\Psi x_i}}_1}_{\text {Interpretation loss }},
\end{equation}

Here the reconstruction loss ensures that the wavelet transform is invertible, allowing for reconstruction of the original data. Hence the transform does not lose any information in the input data.

The wavelet loss ensures that the learned filters yield a valid wavelet transform. Specifically, \cite{mallat1989theory,meyer1992wavelets} characterize the sufficient and necessary conditions on $h$ and $g$ to build an orthogonal wavelet basis. Roughly speaking, these conditions state that in the frequency domain the mass of the lowpass filer $h$ is concentrated on the range of low frequencies while the highpass filter $g$ contains more mass in the high frequencies. We also desire the learned wavelet to provide sparse representations so we add the $\ell_1$ norm penalty on the wavelet coefficients. Combining all these conditions via regularization terms, we
define the wavelet loss at the data point $x_i$ as 
\begin{multline*}
    W(h,g,x_i;\lambda) = \lambda \norm{\Psi x_i}_1 +  (\sum_n h[n]-\sqrt{2})^2 + (\sum_n g[n])^2 + (\norm{h}_2^2-1)^2 \nonumber\\
    + \sum_{w}(|\widehat{h}(w)|^{2}+|\widehat{h}(w+\pi)|^{2} - 2)^2 + \sum_k(\sum_{n} h[n] h[n-2k]-\ones_{k=0})^2,
\end{multline*}
where $g$ is set as $g[n]=(-1)^n h[N-1-n]$ and where $N$ is the support size of $h$ (see~\cite{ha2021adaptive} for further details on the formulations of wavelet loss).

Finally, the interpretation loss enables the distillation of knowledge from the pre-trained model $f$ into the wavelet model. It ensures that attributions in the space of wavelet coefficients $\Psi x_i$ are sparse, where the attributions of wavelet coefficients is calculated by TRIM, as described in \cref{subsec:trim}. This forces the wavelet transform to produce representations that concisely explain the model's predictions at different scales and locations. 

A key difference between AWD and existing wavelet techniques (e.g.~\cite{recoskie2018learning,recoskie2018learningthesis} is that they use \textit{interpretations from a trained model} to learn the wavelets; this incorporates information not just about the signal but also an outcome of interest and the inductive biases learned by a DNN. This can help learn an interpretable representation that is well-suited to efficient computation and effective prediction.

\subsubsection{Reality check: molecular partner prediction}
For evaluation, see \cref{subsec:molecular_prediction}, which shows an example of how a distilled AWD model can provide a simpler, more interpretable model while improving  prediction accuracy.

\section{Real-data problems showcasing interpretations}
\label{sec:real_data_problems}

In this section, we focus on three real-data problems where the methods introduced in \cref{sec:intro_acd} and \cref{sec:improve_models_ovw} are able to provide useful interpretations in context. \cref{subsec:molecular_prediction} describes how AWD can distill DNNs used in cell biology, \cref{subsec:cosmo} describes how TRIM + CD yield insights in a cosmological context, and \cref{subsec:isic_results} describes how CDEP can be used to ignore spurious correlations in a medical imaging task.
\subsection{Molecular partner prediction}
\label{subsec:molecular_prediction}

We now turn our attention to a crucial question in cell biology: understanding clathrin-mediated endocytosis (CME) \cite{kirchhausen2014molecular,he2020dynamics}. It is the primary pathway by which things are transported into the cell, making it essential functions of higher eukaryotic life~\cite{mcmahon2011molecular}. Many questions about this process remain unanswered, prompting  a line of studies aiming to better understand this process \cite{kaksonen2018mechanisms}. One major challenge with analysis of CME, is the ability to readily distinguish between abortive coats (ACs) and successful clathrin-coated pits (CCPs). Doing so enables an understanding of what mechanisms allow for successful endocytosis. This is a challenging problem where DNNs have recently been shown to outperform classical statistical and ML methods.

\begin{figure}[H]
    \centering
    \includegraphics[width=\textwidth]{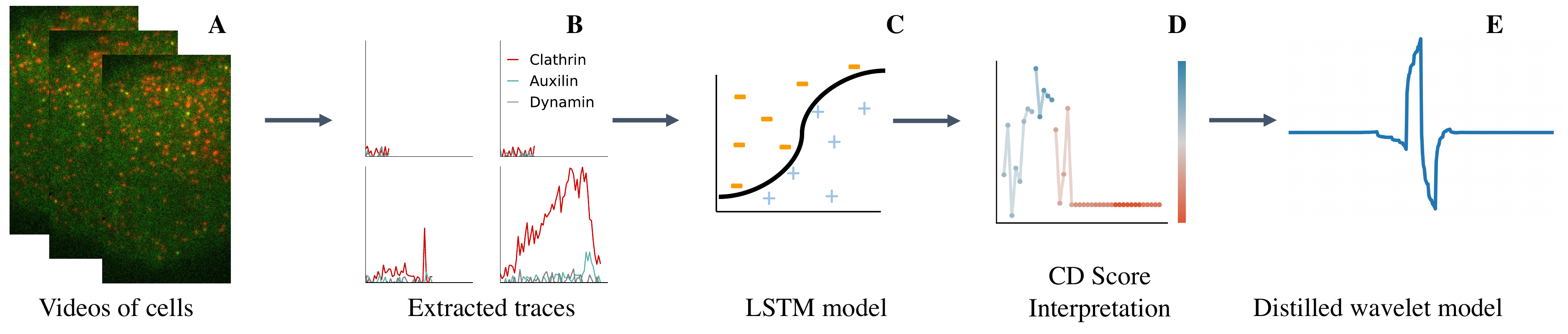}
    \caption{Molecular partner prediction pipeline. \bfp{A} Tracking algorithms run on videos of cells identify \bfp{B} time-series traces of endocytic events. \bfp{C} An LSTM model learns to classify which endocytic events are successful and \bfp{D} CD scores identify which parts of the traces the model uses. \bfp{E} AWD distills the LSTM model into a simple wavelet model which is able to obtain strong predictive performance.}
    \label{fig:aux_pipeline}
\end{figure}

\cref{fig:aux_pipeline} shows the pipeline for this challenging problem. Tracking algorithms run on videos of cells identify time-series traces of endocytic events. An LSTM model learns to classify which endocytic events are successful and CD scores identify which parts of the traces the model uses. Using these CD scores, domain experts are able to validate that the model does, in fact use reasonable features such as the max value of the time-series traces and the length of the trace.

However, the LSTM model is still relatively difficult to understand and computationally intensive.
To create an extremely transparent model, we extract only the maximum $6$ wavelet coefficients at each scale.
By taking the maximum coefficients, these features are expected to be invariant to the specific locations where a CME event occurs in the input data.
This results in a final model with $30$ coefficients ($6$ wavelet coefficients at $5$ scales).
These wavelet coefficients are used to train a linear model, and the best hyperparameters are selected via cross-validation on the training set.
\cref{fig:aux_pipeline} shows the best learned wavelet (for one particular run) extracted by AWD corresponding to the setting of hyperparameters $\lambda=0.005$ and $\gamma=0.043$. \cref{tab:auxilin_results} compares the results for AWD~to the original LSTM and the initialized, non-adaptive DB5 wavelet model, where the performance is measured via a standard $R^2$ score, a proportion of variance in the response that is explained by the model.
The AWD model not only closes the gap between the standard wavelet model (DB5) and the neural network, it considerably improves the LSTM's performance (a 10\% increase of $R^2$ score). Moreover, we calculate the compression rates of the AWD wavelet and DB5---these rates measure the proportion of wavelet coefficients in the test set, in which the magnitude and the attributions are both above $10^{-3}$. The AWD wavelet exhibits much better compression than DB5 (an 18\% reduction), showing the ability of AWD to simultaneously provide sparse representations and explain the LSTM's predictions concisely. The AWD model also dramatically decreases the computation time at test time, a more than $200$-fold reduction when compared to LSTM.

In addition to improving prediction accuracy, AWD enables
domain experts to vet their experimental pipelines by making them more transparent.
By inspecting the learned wavelet, AWD allows for checking what clathrin signatures signal a successful CME event; it indicates that the distilled wavelet aims to identify a large buildup in clathrin fluorescence (corresponding to the building of a clathrin-coated pit) followed by a sharp drop in clathrin fluorescence (corresponding to the rapid deconstruction of the pit).
This domain knowledge is extracted from the pre-trained LSTM model by AWD~using only the saliency interpretations in the wavelet space. 

\begin{table}[H]
    \centering
    \caption{Performance comparisons for different models in molecular-partner prediction. AWD substantially improves predictive accuracy, compression rate, and computation time on the test set. A higher $R^2$ score, and lower compression factor, and lower computation time indicate better results. For AWD, values are averaged over $5$ different random seeds.}
    
\begin{tabular}{lrrr}
\toprule
{} &  \textbf{AWD (Ours)} &  Standard Wavelet (DB5) &   LSTM \\
\midrule
Regression ($R^2$ score)   &     \textbf{0.262 (0.001)} &          0.197  &      0.237 \\
Compression factor            &       \textbf{0.574 (0.010)}     &     0.704        &     N/A\\
Computation time        &      \textbf{0.0002s} &    \textbf{0.0002s}  &    0.0449s\\
\bottomrule
\end{tabular}

    \label{tab:auxilin_results}
\end{table}

To see the effect of interpretation loss on learning the wavelet transforms and increased performance, we also learn the wavelet transform while setting the interpreration loss to be zero. In this case, the best regression $R^2$ score selected via cross-validation is 0.231, and the adaptive wavelets without the interpretation loss still outperforms the baseline wavelet but fail to outperform the neural network models.

\subsection{Cosmological parameter prediction}
\label{subsec:cosmo}
We now turn to a cosmology example, where attributing importance to transformations helps understand cosmological models in a more meaningful feature space. 
Specifically, we consider weak gravitational lensing convergence maps, i.e. maps of the mass distribution in the Universe integrated up to a certain distance from the observer. In a cosmological experiment (e.g. a galaxy survey), these mass maps are obtained by measuring the distortion of distant galaxies caused by the deflection of light by the mass between the galaxy and the observer~\cite{Bartelmann2001}. These maps contain a wealth of physical information of interest to cosmologists, such as the total matter density in the universe, $\Omega_m$.
Current research aims at identifying the most informative features in these maps for inferring the true cosmological parameters, with DNN-based inference methods often obtaining state-of-the-art results~\cite{Zorilla2019,ribli2019improved,Fluri2019}.

In this context, it is important to not only have a DNN that predicts well, but also understand what it learns. Knowing which features are important provides deeper understanding and can be used to design optimal experiments or analysis methods. Moreover, because this DNN is trained on numerical simulations (realizations of the Universe with different cosmological parameters), it is important to validate that it uses physical features rather than latching on to numerical artifacts in the simulations. TRIM can help understand and validate that the DNN learns appropriate physical features by analyzing attributing importance in the spectral domain.

\begin{figure}[H]
    \centering
    \includegraphics[height=1.3in]{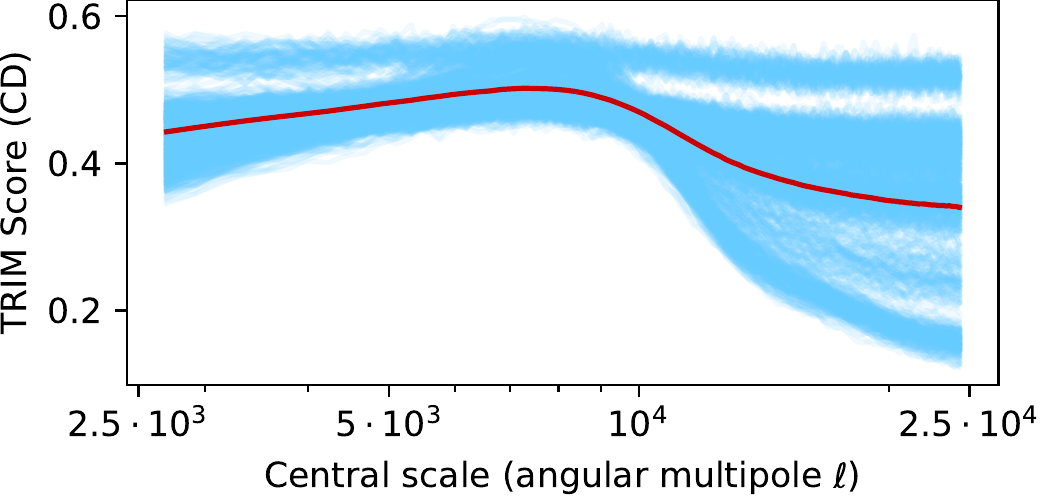}
    \includegraphics[height=1.22in]{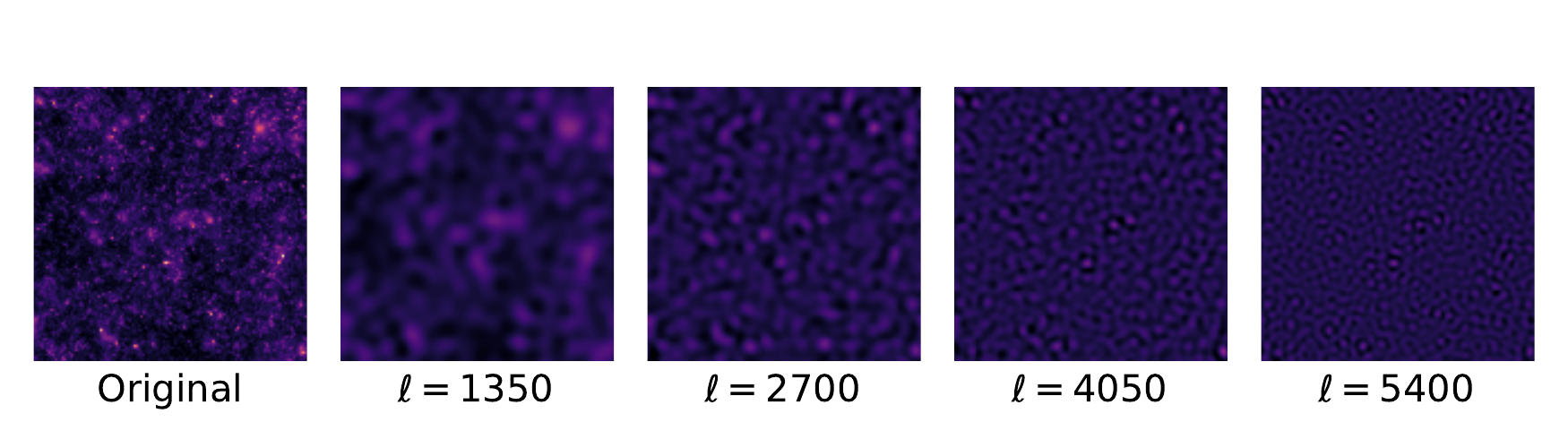}
    \caption{Different scales (i.e. frequency bands) contribute differently to the prediction of $\Omega_m$. Each blue line corresponds to one testing image and the red line shows the mean. Images show the features present at different scales. The bandwidth is $\Delta_\ell=$2,700.}
    \label{fig:fft}
\end{figure}

A DNN is trained to accurately predict $\Omega_m$ from simulated weak gravitational lensing convergence maps (full details in \cite{singh2020transformation}). To understand what features the model is using, we desire an interpretation in the space of the power spectrum. The images in \cref{fig:fft} show how different information is contained within different frequency bands in the mass maps. The plot in \cref{fig:fft} shows the TRIM attributions with CD (normalized by the predicted value) for different frequency bands when predicting the parameter $\Omega_m$. Interestingly, the most important frequency band for the predictions seems to peak at scales around $\ell=10^4$
and then decay for higher frequencies.\footnote{Here the unit of frequency used is angular multipole $\ell$.} A physical interpretation of this result is that the DNN  concentrates on the most discriminative part of the Power Spectrum, i.e. at scales large enough not to be dominated by sample variance, and smaller than the frequency cutoff at which the simulations lose power due to resolution effects.

\begin{figure}[H]
    \centering
    \includegraphics[width=\textwidth]{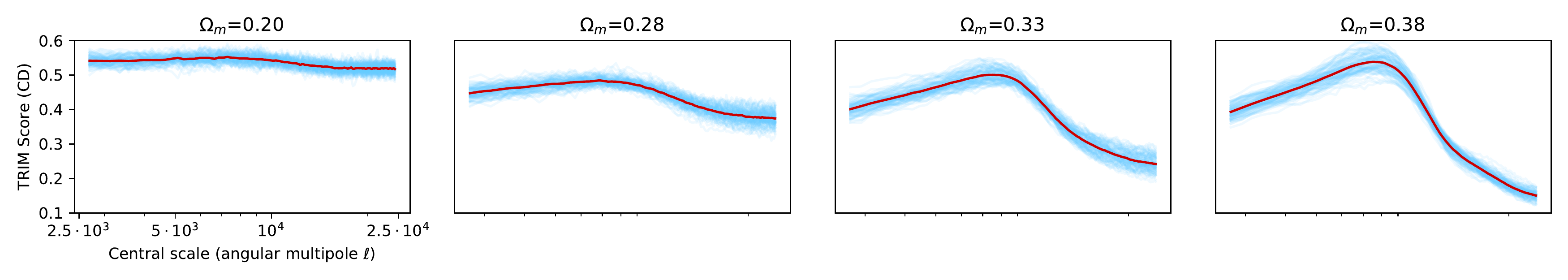}
    \caption{TRIM attributions vary with the value of $\Omega_m$.}
    \label{fig:vary_omegam}
\end{figure}

\cref{fig:vary_omegam} shows some of the curves from \cref{fig:fft} separated based on their cosmology, to show how the curves vary with the value of $\Omega_m$. Increasing the value of $\Omega_m$ increases the contribution of scales close to $\ell=10^4$, making other frequencies relatively unimportant. This seems to correspond to known cosmological knowledge, as these scales seem to correspond to galaxy clusters in the mass maps, which are structures very sensitive to the value of $\Omega_m$.
The fact that the importance of these features varies with $\Omega_m$ would seem to indicate that at lower $\Omega_m$ the model is using a different source of information, not located at any single scale, for making its prediction.

\subsection{Improving skin cancer classification via CDEP}
\label{subsec:isic_results}

\begin{figure}[H]
	\begin{center}
		\includegraphics[width=\textwidth]{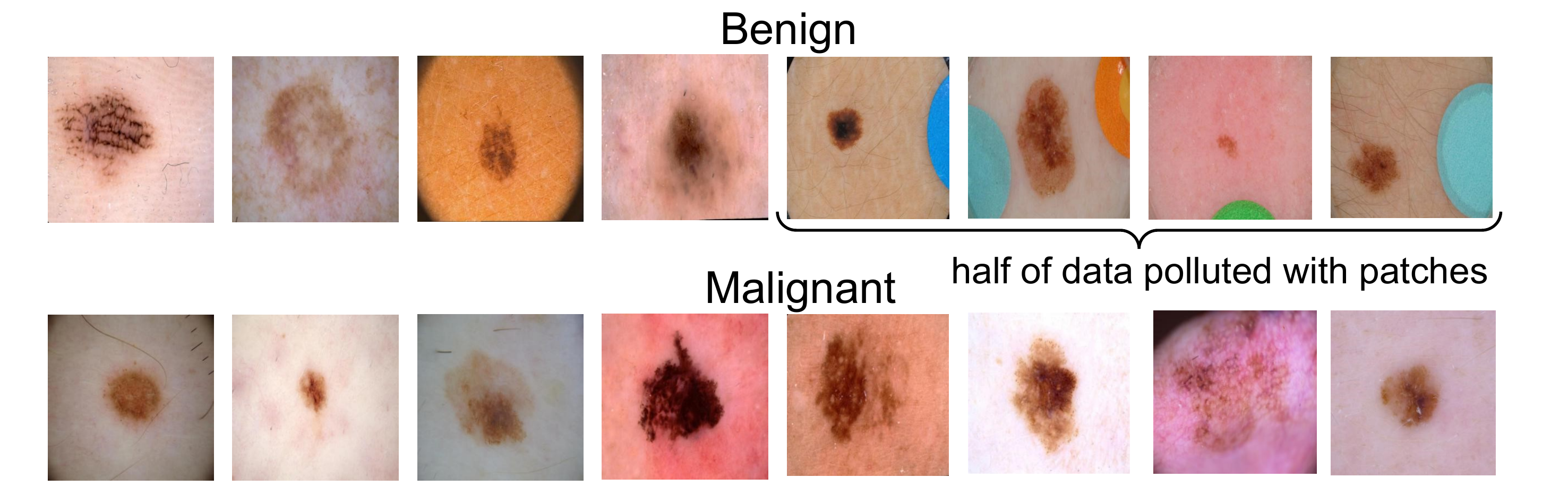}
	\end{center}
	\caption{Example images from the ISIC dataset. Half of the benign lesion images include a patch in the image. Training on this data results in the neural network overly relying on the patches to classify images; CDEP avoids this.
	}
	\label{fig:example_ISIC}
\end{figure}

In recent years, deep learning has achieved impressive results in diagnosing skin cancer~\cite{esteva2017dermatologist}. 
However, the datasets used to train these models often include spurious features which make it possible to attain high test accuracy without learning the underlying phenomena \cite{winkler2019association}. In particular, a popular dataset from ISIC (International Skin Imaging Collaboration) has colorful patches present in approximately 50\% of the non-cancerous images but not in the cancerous images as can be seen in \cref{fig:example_ISIC} \cite{codella2019skin}.
We use CDEP to remedy this problem by penalizing the DNN placing importance on the patches during training.

The task in this section is to classify whether an image of a skin lesion contains (1) benign melanoma or (2) malignant melanoma. In a real-life task, this would for example be done to determine whether a biopsy should be taken.
In order to identify the spurious patches, binary maps of the patches for the skin cancer task are segmented using SLIC, a common image-segmentation algorithm \cite{achanta2012slic}. 
After the spurious patches were identified, they are penalized using to have zero importance.

\cref{tab:ISIC_results} shows results comparing the performance of a DNN trained with and without CDEP. We report results on two variants of the test set. The first, which we refer to as ``no patches'' only contains images of the test set that do not include patches. The second also includes images with those patches. Training with CDEP improves the AUC and F1-score for both test sets, compared to both a Vanilla DNN and using the RRR method introduced in \cite{ross2017right}. Further visual inspection shows that the DNN attributes low importance to regions in the images with patches.

\begin{table}[H]
	\caption{Results from training a DNN on ISIC to recognize skin cancer (averaged over three runs). Results shown for the entire test set and for only the test-set images that do not include patches (``no patches''). The network trained with CDEP generalizes better, getting higher AUC and F1 on both.} 
	\label{tab:ISIC_results}
	\begin{center}
    \begin{tabular}{lrrrrr}
    \toprule
    {} & AUC (no patches) & F1 (no patches) & AUC (all) & F1 (all) \\
    & & & & & \\
    \midrule
    Vanilla & 0.93 & 0.67 & 0.96 & 0.67 \\
    RRR & {0.76 }& {0.45} & {0.87} & {0.45} \\
    CDEP & \textbf{0.95 }& \textbf{0.73} & \textbf{ 0.97} & \textbf{0.73} \\
    \bottomrule
    \end{tabular}
	\end{center}
\end{table}

\section{Discussion} 


Overall, the interpretation methods here are shown to (1) accurately recover known importances for features / feature interactions \cite{murdoch2018beyond}, (2) correctly inform human decision-making and be robust to adversarial perturbations \cite{singh2018hierarchical}, and (3) reliably alter a neural network's predictions when regularized appropriately \cite{rieger2020interpretations}. For each case, we demonstrated the use of reality checks through predictive accuracy (the most common form of reality check) or through domain knowledge which is relevant to a particular domain/audience.

There is considerable future work to do in developing and evaluating attributions, particularly in distilling/building interpretable models for real-world domains and understanding how to better make useful interpretation methods. Below we discuss them in turn.

\subsection{Building/distilling accurate and interpretable models}

In the ideal case, a practitioner can develop a simple model to make their predictions, ensuring interpretability by obviating the need for post-hoc interpretation.
Interpretable models tend to be faster, more computationally efficient, and smaller than their DNN counterparts.
Moreover, interpretable models allow for easier inspection of knowledge extracted from the learned models and make reality checks more transparent.
AWD~\cite{ha2021adaptive} represents one effort to use attributions to distill DNNs into an interpretable wavelet model, but the general idea can go much further.
There are a variety of interpretable models, such as rule-based models~\cite{letham2015interpretable,singh2021imodels} or additive models~\cite{caruana2015intelligible} whose fitting process could benefit from accurate attributions.
Moreover, AWD and related techniques could be extended beyond the current setting to unsupervised/reinforcement learning settings or to incorporate multiple layers.
Alternatively, attributions can be used as feature engineering tools, to help build simpler, more interpretable models.
More useful features can help enable better exploratory data analysis,  unsupervised learning, or reality checks.

\subsection{Making interpretations useful}

Furthermore, there is much work remaining to improve the relevancy of interpretations for a particular audience/problem.
Given the abundance of possible interpretations, it is particularly easy for researchers to propose novel methods which do not truly solve any real-world problems or fail to faithfully capture some aspects
of reality.
A strong technique to avoid this is to directly test newly introduced methods in solving a domain problem.
Here, we discussed several real-data problems that have benefited from improved interpretations~\ref{sec:real_data_problems}, spanning from cosmology to cell biology.
In instances like this, where interpretations are used directly to solve a domain problem, their relevancy is indisputable and reality checks can be validated through domain knowledge.
A second, less direct, approach is the use of human studies where humans are asked to perform tasks, such as evaluating how much they trust a model's predictions \cite{singh2018hierarchical}.
While challenging to properly construct and perform, these studies are vital to demonstrating that new interpretation methods are, in fact, relevant to any potential practitioners.
We hope the plethora of open problems in various domains such as science, medicine, and public policy can help guide and benefit from improved interpretability going forward.

{
    \footnotesize
	 \bibliographystyle{unsrt}
    \bibliography{references.bib}
}
\end{document}